\documentclass{article}

\usepackage{microtype}
\usepackage{graphicx}
\usepackage{subcaption}
\usepackage{booktabs} 

\usepackage{hyperref}

\usepackage[preprint]{icml2026}

\usepackage{amsmath}
\usepackage{amssymb}
\usepackage{mathtools}
\usepackage{amsthm}

\usepackage[capitalize,noabbrev]{cleveref}

\theoremstyle{plain}
\newtheorem{theorem}{Theorem}[section]
\newtheorem{proposition}[theorem]{Proposition}

\theoremstyle{definition}
\newtheorem{definition}[theorem]{Definition}

\theoremstyle{remark}

\usepackage[textsize=tiny]{todonotes}

\icmltitlerunning{Compositional Open-ended Intelligence}

\begin{document}

\twocolumn[
  \icmltitle{A Compositional Framework for Open-ended Intelligence}

  \icmlsetsymbol{equal}{*}

  \begin{icmlauthorlist}
    \icmlauthor{Ida Momennejad}{xx}
    \icmlauthor{Roberta Raileanu}{yy}
  \end{icmlauthorlist}

  \icmlaffiliation{xx}{Microsoft Research NYC, New York, USA}
  \icmlaffiliation{yy}{Google DeepMind, London, UK}

  \icmlcorrespondingauthor{Ida Momennejad}{ida.momennejad@gmail.com}

  \icmlkeywords{Machine Learning, ICML}

  \vskip 0.3in
]
\printAffiliationsAndNotice{}%

\begin{abstract}
Open-ended intelligence is the capacity to adapt to novel problems and environments that are substantially different from those in training. A mathematics of open-ended intelligence requires two pillars: first, a minimal set of representational primitives (e.g., states, actions) and algorithmic primitives (e.g., nearest neighbor); and second, an acquired compositional grammar for selection, recursion, and branching that produces sequences of operations and recurring motifs. We formalize open-ended intelligence in terms of the compositional closure induced by a finite primitive set $P$ and a set of composition operators $C$. We characterize properties of the induced closure $\mathcal{L}(P,C)$ that support unbounded compositional generation across families of tasks and worlds. The closure of the two pillars yields infinite adaptive responses across a wide range of settings. The mathematics supports complementary research agendas, including evaluation metrics for explanation and interpretability, and novel architectures where compositional generalization is native. We propose next primitive prediction (NPP) as a novel architectural objective, where training encourages the acquisition of reusable algorithmic primitives and their compositional grammar, such that new solutions are generated through recombination. Given such an objective, curriculum learning and self-play can enable lifelong learning, expanding the closure by discovering reusable primitives and transition motifs across settings. We ground the framework through case studies in physics, evolution, and neuroscience.

\end{abstract}

\section{Introduction}

Open-ended intelligence is the capacity to generate solutions to problems that were not anticipated during learning \cite{team2021open}, commonly in games and evolutionary computation \cite{lehman2011abandoning, wang2019poet, clune2019aigas}. However, existing open-ended systems typically learn to handle variation within a single environment or procedurally generated games \cite{team2021open, wang2019poet, wang2020enhanced}, rather than compositional generalization and transfer of capabilities across fundamentally different families of environments. Machine learning lacks a compositional framework for open-ended intelligence, with a mathematics of primitives, composition, and closure.

The dominant framing centers on "stepping stones" \cite{lehman2011abandoning, stanley2017openended, clune2019aigas}, that is, behavioral and skill repertoires that accumulate over time and enable further discovery. This framing captures how complex solutions emerge without a fixed objective and through intermediate discoveries \cite{stanley2017openended}, and some even argue that open-endedness is required for superhuman intelligence \cite{hughes2024openendedness}. However, the stepping-stone framework does not natively capture compositional generalization. In the current framing an agent can reuse a particular skill if it encounters a similar scenario, but this reuse is episodic rather than compositional. The framework does not guarantee efficient learning, nor learning of the required latent representations of algorithmic steps or primitives \cite{lippl2025primitives}, the required inference rules, or their algorithmic compositions.

\begin{figure*}[t]
    \centering
    \includegraphics[width=0.9\textwidth]{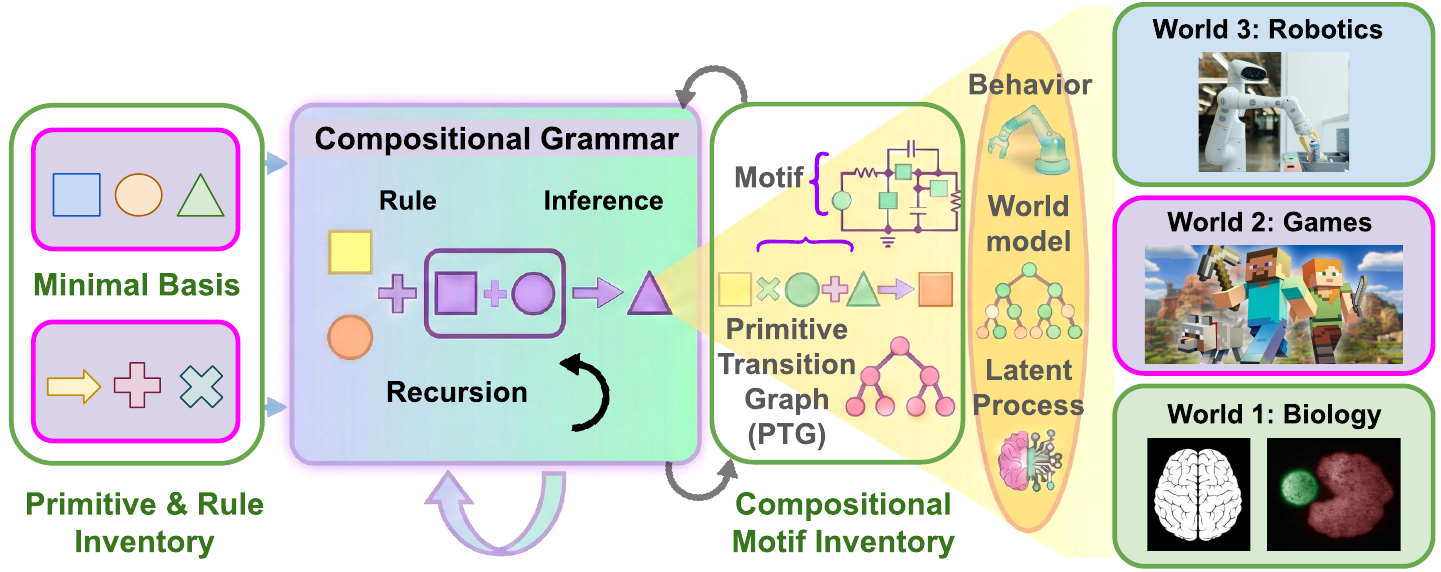}
    \caption{\textbf{Compositional Open-Ended Intelligence.} A minimal basis of primitives and operators feeding into a compositional grammar, which leads to an unbounded closure of increasingly complex solutions across infinite worlds.}
\label{fig:compositional_intelligence}
\end{figure*}

A compositional framework for open-ended intelligence both captures the ontology of open-ended exploration across families of worlds and scales, and enables architectural innovations that make compositional generalization native. Here a primitive is a minimal unit of computation or representation that can be reused across contexts, and has been identified in the latent representations across the layers of LLMs \cite{eberle2025position, lippl2025primitives}. Two types of primitives are central to our framework. \textit{Representational primitives} capture what combinations of features constitute the agent's ontology, i.e., what the system perceives as objects, relations, and parts of the world it operates in. Here, a world-model \cite{ha2018world} is a combination of representational primitives. \textit{Algorithmic primitives} capture what minimal computational or algorithmic operations the system pieces together to compose a solution \cite{eberle2025position, lippl2025primitives}. Examples include nearest-neighbor retrieval, distance computation, verification, and comparison \cite{lippl2025primitives}. Composition refers to how the system pieces the primitives together. It can be captured in terms of \textit{transition motifs} among primitives, including sequential transitions and recurrence, that are traversed in the search for a solution. \textit{Compositional geometry} refers to latent weight patterns corresponding to primitives and composition in the underlying model \cite{eberle2025position}. See Figure~\ref{fig:compositional_intelligence} for a diagram of our framework. 

We define open-ended exploration as a search over the right primitives and the right transition motifs to compose them into a solution. This is distinct from search over actions or skills alone. We hypothesize that the closure of a minimal set of primitives and composition rules yields an unbounded space of algorithms, representations, and behaviors. Current frameworks lack explicit computational and mathematical objects to capture compositional generalization, learning, and reuse \cite{GopnikEtAl2004CausalMapsBayesNets}. Just as in brains, open-ended  exploration and problem solving requires a compositional grammar of planning a sequence of operations (internal actions rather than actions in the outside world) to solve a given problem \cite{Momennejad2026COBS}.

This paper makes four contributions to open-ended intelligence. It introduces a formal object for compositional open-ended intelligence defined by a primitive set \(P\), composition operators \(C\), and the induced closure \(\mathcal{L}(P,C)\). First, we formalize primitives, composition motifs, and closure, and we define open-endedness in terms of formal properties of \(\mathcal{L}(P,C)\). Second, we derive implications for architecture and evaluation (including the Next Primitive Prediction objective and compositional transfer metrics) and illustrate these consequences through mathematical, biological, and neural case studies. We also discuss collective open-ended intelligence, where primitives and compositions are preserved, transmitted, and recombined across agents, populations, and time. The framework unifies the study of natural and artificial open-ended intelligence by focusing on transferable compositional generalization at individual, mechanistic, and collective scales.

\section{Existing Frameworks and Related Works}

Current algorithms for Open-Ended Intelligence typically search in behavioral space ($\mathcal{B}$) or program space ($\Pi$) (see Table~\ref{tab:collapsed_families}). Open-ended discovery is often framed through "stepping stones"~\citep{lehman2011abandoning}, which enable behavioral repertoires to accumulate via co-evolutionary systems such as POET~\citep{wang2019poet, wang2020enhanced} and XLand~\citep{team2021open}. These frameworks often use Quality Diversity (QD) algorithms~\citep{cully2015robots, cully2017quality, samvelyan2023maestro} and Procedural Content Generation (PCG)~\citep{yannakakis2011experience,summerville2018procedural,grbic2021evocraft}. However, their ``reuse" remains largely episodic, agents rarely decompose skills into transferable algorithmic primitives. 

\subsection{Novelty Search}

In Novelty Search (NS)~\citep{lehman2011abandoning}, the algorithm optimizes for novel, rather than fit, behaviors, maximizing  $\Phi(\theta) = \text{dist}(b_{\theta}, \mathcal{B}_{\text{archive}})$, where  $\theta$ are the parameters of the policy and $\mathcal{B}_{\text{archive}}$ is a set of previously discovered behaviors. MAP-Elites algorithms~\citep{mouret2015illuminating} such as Quality-Diversity (QD)~\citep{pugh2016quality} seek to find the highest-performing solution for every possible behavioral niche, so that $\forall N_i, \max Q(\theta)$ subject to $b_{\theta} \in N_i$, where $Q$ is a quality measure. While this allows the system to replay a ``skill'', it cannot decompose it into reusable primitives. However, optimizing for novelty alone is prone to the ``noisy TV" problem, or confusing noise with novelty ~\citep{schmidhuber1991curious, oudeyer2007intrinsic, burda2019exploration, pathak2017curiosity}. To mitigate this, \cite{hughes2024openendedness} argue that an open-ended system should optimize not only for novelty but also learnability. A recently proposed ``epiplexity" measure formalizes the information that bounded observers can extract from data, distinguishing learnable structural content from unpredictable noise \cite{finzi2026epiplexity}.

\subsection{Autocurricula and Skill Accumulation}

Autocurricula approaches \cite{TeohEtAl2025NextLatentPrediction} use co-evolution to generate an endless stream of increasingly difficult environments, such that $\pi_{i+1} = \text{train}(\pi_i, E_{\text{new}})$ and $E_{\text{new}}$ is generated via mutations of successful environments. Progress is a sequence of increasingly more complex behaviors rather than the emergence of a minimal generating set of primitives and rules.

Voyager \cite{wang2023voyager} presents a framework for skill accumulation in interactive worlds, where an embodied agent uses LLMs to build a library of skills in the form of executable code. An LLM-based automatic curriculum proposes subgoals and identifies the skills required to achieve these subgoals, an executor generates the corresponding code for each skill, and a hierarchical policy composes existing skills to achieve the given subgoals. However, there is no guarantee the learned set of skills is minimal and the composition of skills is determined by the LLM's weights rather than a learned formal closure. LLM-guided evolutionary search methods \cite{romera2024mathematical,novikov2025alphaevolve, teodorescu2023codeplay,ma2023eureka} crucially use LLMs as evolutionary operators to evolve code fragments rather than a basis of primitives or transition motifs. 
 

\subsection{Open-ended Exploration, Curiosity, and Stepping Stones} 
The dominant theoretical framing for open-ended discovery centers on the concept of ”stepping stones,” where behavioral and skill repertoires accumulate over time to enable further discovery \cite{lehman2011abandoning}. Systems like POET~\citep{wang2019poet, wang2020enhanced} and XLand~\citep{team2021open} have successfully demonstrated the co-evolution of agents and environments within procedural families. These approaches often build upon Quality Diversity (QD) algorithms, which seek to maintain a collection of high-performing, yet diverse solutions~\citep{cully2015robots, cully2017quality,samvelyan2023maestro}, allowing robots to adapt to damage or new environments much like animals do. This lineage is deeply rooted of Procedural Content Generation (PCG). Early frameworks focused on experience-driven PCG \citep{yannakakis2011experience} to optimize for player affect, which eventually transitioned into PCG via Machine Learning (PCGML) \citep{summerville2018procedural}, where models learn to capture the statistical distributions of human-designed content. Benchmarks like EvoCraft \citep{grbic2021evocraft} push this further, framing the open-ended evolution of complex structures in voxel worlds as a primary challenge for general intelligence. 

However, as we highlight, this form of reuse is often episodic rather than truly compositional; an agent might replay a specific skill if it encounters a similar scenario, but it does not necessarily decompose that skill into transferable algorithmic primitives. Recent advancements in LLM-powered discovery have attempted to bridge this gap. For instance, Voyager \cite{wang2023voyager} uses an automated curriculum and a skill library to facilitate lifelong learning, while FunSearch~\citep{romera2024mathematical}  and Eureka~\citep{ma2023eureka} demonstrate that LLMs can discover new mathematical truths or reward functions by searching the space of verifiable programs. Similarly, AlphaEvolve~\citep{novikov2025alphaevolve} and Codeplay~\citep{teodorescu2023codeplay} explore the automated evolution of model architectures and code-based policies.

Furthermore, exploration driven by Intrinsic Motivation or curiosity-driven developmental processes provides excellent explanation hooks for why discovery proceeds~\citep{schmidhuber1991curious, schmidhuber1997discovering, schmidhuber2010formal}, but often lacks a concrete architectural target for engineering. Here we propose next-primitive prediction as a training objective. Unlike JEPA~\citep{assran2023ijepa} or next-latent prediction \cite{TeohEtAl2025NextLatentPrediction}, which target state/feature dynamics, our objective targets the grammar of computation, forcing the architecture to prioritize a compact, role-flexible basis that remains stable across counterfactual variations of possible worlds. This moves beyond the ”stepping stone” lineage by defining progress as the emergence of reusable primitives and a compositional grammar.

\subsection{Systematic Generalization and Planning}
The challenges of systematic recombination are well-documented in benchmarks like ARC~\citep{chollet2024arc}, which diagnoses failures in human-like abstraction. Previous attempts to address this via planning, such as Composable Planning with Attributes~\citep{zhang2018composable}, utilize attribute-conditioned policies for zero-shot pathfinding. However, these approaches lack a formal pressure toward a minimal generating set—the parsimony constraint required to consolidate knowledge into a compact, reusable basis. This gap is partially addressed by Machado's work on eigenoptions~\citep{machado2017laplacian}, which utilizes the eigenvectors of the graph Laplacian to discover a set of intrinsic options that capture the multi-scale geometry of the state space. While eigenoptions provide a spectral basis for diffusion and exploration, they primarily serve as a representation of the environment's topology rather than a generative grammar for task-specific composition. Unlike Causal Learning~\citep{gopnik2004theory} or SCAN-style critiques that diagnose failures behaviorally~\citep{lake2018generalization}, our framework formalizes the mathematics of basis and closure.

\subsection{Representational and Object-Centric Approaches}
A significant body of work has sought to establish representational primitives through object-centric learning. Architectures such as Slot Attention~\citep{locatello2020object}, MONet~\citep{burgess2019monet}, and IODINE~\citep{greff2019multi} demonstrate the power of factorizing scenes into discrete entities, providing a necessary ontology for world modeling. Similarly, the Core Knowledge perspective~\citep{spelke2007core} identifies innate cognitive priors for objects, agents, and geometry. While these models provide a necessary representational basis, they typically treat the process of composition as an implicit property or a fixed architectural prior. In contrast, our framework elevates algorithmic primitives (i.e., minimal computational operations like verification, retrieval, or comparison) to first-class citizens alongside representational ones. We prioritize the transition motifs (sequencing, branching, recurrence) that stitch these primitives into primitive transition graphs as the primary transferable unit.

\begin{table*}[t]
\caption{Collapsed families of open-ended systems against the proposed compositional framework.}
\label{tab:collapsed_families}
\begin{center}
\begin{scriptsize}
\renewcommand{\arraystretch}{1.25}
\setlength{\tabcolsep}{6pt}
\begin{tabular}{p{2.2cm} p{1.8cm} p{2.3cm} p{4.2cm} p{4.6cm}}
\toprule
Family &
System &
Primary search object &
How ``open-endedness'' is induced &
Core gap w.r.t. primitives and composition \\
\midrule

Novelty Search, Quality-Diversity &
Novelty Search, QD/MAP-Elites &
Behavioral descriptors &
Reward behavioral divergence (or quality across niches) to accumulate repertoires &
Diversity in outcome space, not composition space, reuse is recall not recomposition \\

Autocurricula, Env-Agent Coevolve &
POET, XLand &
Task distributions &
Expand task families, train continually across the growing distribution &
Episodic skill reuse, no primitive decomposition \\

LLM-Guided Evo. Search &
FunSearch, AlphaEvolve &
Code artifacts &
Iterative propose evaluate select over candidate programs &
Code fragments (not minimal basis or composition motif) \\

Interactive LLM Skills &
Voyager &
Skill/tool libraries &
Skills/tools accumulate via iterative exploration &
Skill accumulation (no primitives nor composition rules or motif) \\
\midrule

\textbf{Ours: primitives \& composition closure} &
\textbf{This work} &
\textbf{Primitive set and composition motifs} &
\textbf{Learn/measure a minimal generating set, transfer via recomposition across worlds} &
\textbf{Primitives /operators/ motifs primary objects to align architecture, algorithm, evaluation} \\
\end{tabular}
\end{scriptsize}
\end{center}
\end{table*}

Our framework is motivated by findings in computational neuroscience showing that high-dimensional representations with mixed selectivity support flexible computation and compositional reuse \cite{fusi2016neurons, rigotti2013importance}, and that prefrontal and hippocampal circuits organize abstract variables in disentangled geometries that enable generalization across contexts \cite{bernardi2020geometry, courellis2024abstract}. The geometry of neural representations thus provides an existence proof that biological intelligence solves the compositional generalization problem through structured primitives and their recombination. Moreover, evidence from causal learning in children~\citep{GopnikEtAl2004CausalMapsBayesNets} and building Xenobots from skin cells ~\citep{kriegman2020scalable, blackiston2021cellular, kriegman2021kinematic} offer behavioral and biological existence proof of compositionality. While these systems ground the discovery process, they lack formal compositional closure.

\section{A Compositional Framework for Open-Ended Intelligence}

We propose that true open-ended intelligence is rooted in parsimony constraints, where progress is measured by learning a smaller, more reusable basis and a simultaneous expansion in the diversity and depth of reachable compositions. Open-ended intelligence also requires a theory of what drives continued primitive discovery and compositional expansion when objectives drift or when no fixed objective is adequate. In natural systems, exploration is guided by novelty, surprise, curiosity, and ecological relevance \cite{cisek2010neural}, what is informative or valuable for the organism given its niche \cite{schmidhuber2010formal, OudeyerSmith2016OpenEndedDevelopment, gottlieb2013information}. A compositional framework reframes ``interestingness" in internal terms. What counts as "interesting" is what expands the agent's closure, new primitives that unlock new compositions, and new composition motifs that allow old primitives to be deployed in qualitatively new ways. Evolution provides an existence proof that long-horizon novelty can be organized around reusable building blocks and their recombination under constraints \cite{wagner1996complex, kirschner1998evolvability, carroll2005endless, watson2016can}, rather than around direct optimization of a fixed short-horizon metric \cite{simon1962architecture, stanley2015greatness}.
Thus, a mathematics of open-ended intelligence is needed to capture selection and branching that operate over compositional structure, possibly governed by a meta-learned grammar of compositionality that can be extracted from latent-space operations \cite{Momennejad2026COBS}.

\subsection{Compositional Open-ended Intelligence}

We redefine open-ended intelligence as the capacity to acquire a library of primitives and a compositional grammar and motifs that generalize across diverse families of environments. In this framework, ``stepping stones,'' are not behavioral episodes but reusable nodes and motifs within a primitive transition graph.

\begin{definition} \textit{Representational primitives} capture the ontology or world model, an agent's perception of objects and states. \textit{Algorithmic primitive} refers to a minimal computational operation observed in a reasoning process, such as comparison or retrieval. \textit{Algorithmic composition} refers to how the primitives are pieced together to form a solution to a problem. Composition is formalized as  transition motifs that rearrange primitives into a world model or strategy, where (a) \textit{compositional grammar} captures rules like sequencing and recurrence for recombination and inference, and (b) \textit{compositional motif} captures common transition structures or algorithmic "phrases" that can be represented as a graph motif, e.g., castling in chess or the hero's journey story structure.
\end{definition}

\begin{definition}Open-endedness is defined as the capacity to search over the closure $\mathcal{L}(P,C)$ induced by a minimal set of primitives $P$ and composition operators $C$, where the system generates an unbounded space of algorithms, representations, behaviors, and solutions across worlds. 
\end{definition}

Crucially, requiring or enforcing a minimal basis to cover an unbounded space of solutions (e.g., via contrastive learning \cite{hadsell2006dimensionality, oord2018representation, chen2020simple, he2020momentum}) forces the system to represent worlds as reusable parts and composition rules rather than as stored whole solutions. This can be done by contrastive learning. Contrastive learning objectives, which encourage representations to bring similar inputs closer and push dissimilar inputs apart, originate in metric learning approaches \cite{hadsell2006dimensionality} and were extended to self-supervised representation learning via contrastive predictive coding \cite{oord2018representation} and modern large-scale frameworks such as SimCLR and MoCo \cite{chen2020simple, he2020momentum}. The idea here, which we expand in the architecture proposal later, is that the same method can be used to reward the model to reuse operations. Once learned, a minimal set of primitives and their compositional grammar enables generalization to novel worlds via re-composition.

\textbf{Primitive hierarchy.} Compositional motifs naturally lead to a hierarchy of primitives and motifs across scales and contexts. Consider "walking" behavior, its underlying algorithmic primitives may include distance computation or balance verification in one world, and control of robotics in another. Walking emerges as a motif, multiply realizable across different contexts with the relevant primitives. 

\subsection{Vector-based Steering and Control of Primitive-level Objects}
A growing line of work suggests that reusable computational units can be operationalized as manipulable directions in a model's latent space. Function vectors show that specific transformations can be reliably induced by applying learned directions, providing an empirical handle on what we term algorithmic primitives: minimal, reusable units of computation \cite{todd2024function}. Persona vectors extend this to broader behavioral profiles, which can be monitored and steered via latent directions \cite{chen2025personavectorsmonitoringcontrolling}; in our framework, such profiles correspond to context-dependent priors over primitive selection and over the distribution of transition motifs. Finally, recent findings that the latent representations of LLMs during reasoning decomposes into algorithmic primitives and their compositional geometry \cite{lippl2025primitives}. Vector-based control offers evidence for primitive-level objects and their compositions as candidate computational objects that can be causally extracted, induced, and evaluated.

\begin{table*}
\centering
\caption{Extended Comparison of Open-Ended Intelligence Frameworks}
\label{tab:oei-comparison}
\resizebox{\textwidth}{!}{%
\begin{tabular}{@{}p{3.2cm}p{3.5cm}p{4cm}p{4.5cm}p{5cm}@{}}
\toprule
\textbf{Paper/System} & \textbf{Objective} & \textbf{What ``open-endedness'' means} & \textbf{Compositional Generalization} & \textbf{What's Missing} \\
\midrule

Evolutionary Algorithms & 
Optimization via selection, mutation, crossover & 
Population-based search inspired by natural selection & 
Implicit via genetic recombination; no explicit primitive inventory or composition rules & 
No explicit primitives, no composition operators, no closure properties. Recombination is random, not structured by inference rules. \\
\midrule

POET \cite{wang2019poet} & 
Endlessly generate increasingly complex environments and solutions & 
Co-evolution of agent-environment pairs; stepping stones enable new challenges & 
Stepping stones are behavioral, not compositional. No primitive reuse or role-flexible binding. & 
Stepping stones are episodic behaviors, not reusable primitives. No transition motifs, no minimal generating set. \\
\midrule

Enhanced POET \cite{wang2020enhanced} & 
POET with domain-general novelty measure & 
Unbounded invention of challenges; meaningful novelty detection & 
Novelty measured behaviorally, not at level of compositional structure. & 
Novelty is behavioral diversity, not compositional expansion. No consolidation toward minimal basis. \\
\midrule

XLand \cite{team2021open} & 
Train agents general across vast task space & 
Iterative improvement across procedurally generated 3D games & 
Implicit via multi-task training; no explicit primitive discovery or transfer-as-recomposition. & 
Generalization is statistical, not compositional. No primitive transition graphs, no motif learning. \\
\midrule

AdA \cite{bauer2023human} & 
Human-timescale in-context adaptation & 
Meta-RL; adaptation without weight updates & 
Adaptation is to task variants, not compositional recombination of primitives. & 
Adaptation is episodic hypothesis testing, not primitive selection. No composition rules or transition structure. \\
\midrule

FunSearch \cite{romera2024mathematical} & 
Discover mathematical knowledge via program search & 
LLM + evolutionary algorithm to evolve functions & 
Programs are compositions, but search is mutation + selection, not learned composition rules. & 
No explicit primitive inventory. Composition operators not learnable objects. Closure implicit in program space. \\
\midrule

AlphaEvolve \cite{novikov2025alphaevolve} & 
General-purpose algorithm discovery & 
LLM ensemble + evolution to evolve codebases & 
Evolves code compositionally but primitives are code fragments, not learned minimal basis. & 
No minimal generating set discovery. No transfer-as-recomposition across domains. \\
\midrule

Open-Endedness for ASI \cite{hughes2024openendedness} & 
Open-endedness as essential for ASI & 
Artifacts that are novel and learnable to an observer & 
Interestingness defined by observer learnability, not compositional structure. & 
Novelty/learnability are outcome metrics. No primitives, no composition rules, no closure. \\
\midrule

Novelty Search \cite{lehman2011abandoning} & 
Abandon objectives; search for novelty & 
Divergent search without fixed fitness function & 
No compositional structure. Novelty is behavioral distance in outcome space. & 
Novelty measured in behavior space, not composition space. No primitive discovery or transition motifs. \\
\midrule

JEPA \cite{assran2023ijepa} & 
Predict next latent state/features & 
Self-supervised learning via latent prediction & 
Learns representations but not explicit primitives or composition rules. & 
Predicts latent dynamics, not primitive compositions. No algorithmic primitives, no grammar of computation. \\
\midrule

\textbf{Compositional Open-ended Intelligence (COI) here} & 
Next-primitive prediction; learn minimal basis + composition rules & 
Search over primitives and transition motifs whose closure yields unbounded exploration & 
\textbf{Native}: Explicit primitives, composition operators, transition motifs, closure, minimal generating set, transfer-as-recomposition. & 
Addresses missing mathematics: primitives, composition rules, closure, parsimony, and compositional evaluation metrics (PRI, CDG, TaR). \\

\bottomrule
\end{tabular}%
}
\end{table*}

\subsection{Compositional Transfer over Possible Worlds}

In philosophy, possible worlds are a tool for reasoning about necessity and contingency. Leibniz introduced the idea of families of possible worlds distinguished by what is compossible \cite{leibniz1710theodicy}, and the framework was later formalized for counterfactual reasoning by Stalnaker, Lewis, and Kripke \cite{stalnaker1968theory, lewis1973counterfactuals, lewis1986plurality, kripke1980naming}. They consider variants of the actual world that differ in specific, controlled ways and ask what holds across them. Epistemologists use the notion of a neighborhood of nearby possible worlds to distinguish genuine knowledge from lucky guesses \cite{nozick1981philosophical, pritchard2005epistemic}. Here we define a \textbf{task family} as a neighborhood of possible worlds, and compositional generalization in terms of what survives across them. Primitives and transition motifs that persist when you vary dynamics, swap reward structures, or reassign which objects play which roles are candidates for genuine open-ended compositional generalization rather than artifacts of a particular world.

In this context, open-endedness requires the capacity to transfer knowledge to novel worlds by decomposing tasks into components observed across different families. This fundamental transfer problem is addressed by deriving a minimal set of algorithmic primitives and composition rules, which are formalized as transition graph motifs, through induction and ablation logic. Ultimately, open-ended intelligence emerges from the strategic recombination and transfer of these primitives and motifs across various possible worlds. Open-endedness is therefore not primarily a property of a behavioral repertoire; it is a property of a transition structure over primitives. A solution in this framework is described as a primitive transition graph: which primitives are invoked, in what order, with what branching and recurrence, and under what conditions. Stepping stones, thus, become visible as reusable nodes, subgraphs, and motifs in this transition graph, rather than as behavioral episodes that are simply replayed. An algorithm can be represented as a traversal through primitives with branching and recurrence, modeled as a graph or a Markov transition process.

\subsection{Evaluation, Interpretability, Architecture.}
The framework serves three primary purposes: it enables mechanistic, representational, and algorithmic interpretability to evaluate internal reasoning \cite{lippl2025primitives, zou2025representationengineeringtopdownapproach}, it supports architectures like ``next-primitive prediction" where compositional generalization is a native property, and it facilitates \textbf{collective open-ended intelligence} through the exchange of portable, functional primitives across agents. Just as next-token prediction in transformer architectures led to the learning of the grammar of language, next-primitive prediction, where a primitive is defined both in behavior and in the latent space, could lead to the discovery of the compositional grammar of open-ended intelligence. 

\section{Mathematics of Open-Ended Intelligence}
Here we propose a mathematics of basis and closure for open-ended intelligence. Let $P$ denote a set of primitives (representational and algorithmic) and let $C$ denote a set of composition/inference operators that specify how primitives can be combined into larger computations. The closure induced by $(P, C)$ is the set of computations, representations, and strategies reachable by repeated application of $C$ to elements of $P$. Open-endedness is defined by whether the induced closure supports unbounded generation of novel compositions within a family of worlds. Closure only becomes meaningful if it includes a parsimony constraint, as without pressure toward a minimal generating set, a system can expand its repertoire by accumulating brittle, task-specific fragments. The minimal generating set problem is therefore central: what is the smallest set of primitives and operators whose closure supports open-ended intelligence? 

\subsection{The Core Tuple}

A compositional agent is defined by the triple $(\mathcal{P}, \mathcal{C}, \mathcal{L})$. 

\textit{$\mathcal{P}$ (Primitive repertoire or library):} A finite set of representational primitives (axioms, fields, objects) and algorithmic primitives (comparison, retrieval, verification). 

\textit{$\mathcal{C}$ (Composition Operators):} A set of motifs that chain primitives, such as sequencing, recursion, and branching. 

\textit{$\mathcal{L}(\mathcal{P}, \mathcal{C})$ (Closure):} The set of all computations reachable by repeatedly applying $\mathcal{C}$ to $\mathcal{P}$. Open-endedness is defined by whether $|\mathcal{L}(\mathcal{P}, \mathcal{C})| = \infty$.

\begin{proposition} [Unbounded compositional closure] Let \(P\) be a finite set of primitives and \(C\) a set of composition operators that includes at least one recursive or generative operator. If compositions under \(C\) preserve type-consistency and admit reuse of intermediate outputs, then the induced closure \(\mathcal{L}(P, C)\) contains an unbounded number of distinct compositions.

\end{proposition}

We use closure in its standard algebraic sense: the closure L(P, C) of a primitive set P under composition operators C is the set of all computations, representations, and strategies reachable by repeatedly applying C to elements of P. The natural numbers — the closure of {0} under the successor function — illustrate the key point: an infinite set generated by a single compact rule. Open-endedness in our sense is the property that a minimal generating set yields an infinite, useful closure; the size of the closure alone is not the criterion, since an unconstrained basis produces an infinite one just as easily. 
Verification primitives and the description-length pressure of Section 4 bound which compositions enter the effective closure, and the substantive problem the framework poses is the minimal generating set problem: the smallest P and C whose closure covers the relevant space of tasks.

\subsection{The Primitive Transition Graph (PTG)} Every solution can be represented as a directed graph $G = (P, C)$ where the nodes are primitives $p \in \mathcal{P}$ invoked during the solution, and edges are instances of operators $c \in \mathcal{C}$ transforming states between primitives. Recurring subgraph motifs are preserved and transferred across task families (Figure 1). The emergence of structured Primitive Transition Graph (PTG) \cite{lippl2025primitives} is governed by a tripartite process of extraction, consolidation, and predictive optimization. This loop ensures that the agent does not merely memorize sequences but identifies the underlying generative grammar of the task. 

\subsection{The Parsimony Constraint} Selection is guided by a Minimum Description Length (MDL) objective. The algorithm asks "What is the smallest set of functions $\mathcal{P}$ that can reconstruct all successful traces in the training set?" Operators ($\mathcal{C}$) are learned as the "edges" that satisfy type-safety between primitives. If primitive $A$ outputs a tensor and primitive $B$ requires a tensor, "Sequencing" is a valid candidate operator.

\subsection{Discovering Motifs} Motifs are ``meaningful phrases" in the language of compositional computation. They can emerge or be discovered via sub-graph mining over PTGs of many different successful solutions. For instance, if the sequence [Calculate Distance] $\rightarrow$ [Threshold Check] $\rightarrow$ [Branch] appears in 90\% of navigation tasks, it can be consolidated into a motif. Once a motif is identified, it can be "wrapped" and becomes a higher-order primitive or operation in the inventory, yielding a hierarchy. This is similar to turning a recurring block of code into a reusable function.

\textbf{Example 1. Algorithm Discovery.} Our framework replaces the mutation of brittle code fragments (ASTs) with a search for the optimal Primitive Transition Graph (PTG) that exists within a formal closure. By utilizing representational primitives ($\mathcal{P}_{\text{rep}}$) like tensors and indices alongside algorithmic primitives ($\mathcal{P}_{\text{alg}}$) such as recursive partitioning, the system avoids brittle solutions common in systems like FunSearch~\citep{romera2024mathematical}. While a standard code-evolver might discover an optimization for a specific input size that fails on the next, a compositional approach learns the underlying generating rules through a parsimony constraint, ensuring the algorithm remains flexible.

\textbf{Example 2. Open-World Exploration} For embodied interaction in open-ended worlds like Minecraft, the framework moves beyond the accumulation of ``black-box" skills found in agents like Voyager~\citep{wang2023voyager}. Instead of treating a skill as an opaque function, the system decomposes behavior into a shared pool of re-bindable primitives and transition motifs, such as $\text{Search} \to \text{Interact} \to \text{Verify}$. Through Transfer-as-Recomposition (TaR), an agent encountering a novel environment, such as an underwater world, can simply re-bind its existing $move\_to$ primitive to $swim\_to$ while keeping the high-level transition logic identical. This architectural approach ensures that competence survives across counterfactual variations of possible worlds rather than being contingent on a single fixed setup.

\section{Architecture: Next Primitive Prediction}

Developing a specialized architecture for open-ended intelligence requires moving beyond the limitations of next-token, next-state, and next-latent prediction. While approaches like JEPA \cite{assran2023ijepa} predict state and feature dynamics, they do not center  primitive inventories or composition rules. We propose 
Next Primitive Prediction (NPP) as a composition-native architectural objective designed to induce a formal grammar of computation. Just as next-token prediction induces linguistic grammar, NPP forces the discovery of a compact, role-flexible basis that remains stable across diverse world families. Here open-endedness arises from the strategic recombination of a minimal generating set rather than the pursuit of scale or reward shaping.

\begin{definition}[Next Primitive Prediction]
Let $\mathcal{P}$ be a finite set of primitives and $\mathcal{C}$ a set of composition operators. Given a partial traversal $(p_{1:t}, c_{1:t})$ of a primitive transition graph in a world $W$, Next Primitive Prediction (NPP) is the objective of predicting the next primitive--operator pair
\begin{equation}
\begin{aligned}
(p_{t+1}, c_{t+1}) &\sim p_\theta(\cdot \mid p_{1:t}, c_{1:t}, W),\\
&\quad (p_{t+1}, c_{t+1}) \in \mathcal{P} \times \mathcal{C}
\end{aligned}
\end{equation}
Here, the training objective favors reuse and recombination of primitives $P$ with their compositional grammar $C$. New solutions emerge from recomposition of existing primitives rather than from task-specific fragments or new bespoke operations.
\end{definition}

\subsection{Training the Next Primitive Predictor (NPP).} The core of the generative architecture is the Next Primitive Predictor (NPP), typically implemented as a transformer-based graph neural network. The training process begins with input encoding, where the current state of the environment ($S_t$) and the cumulative history of the graph ($G_t$) are mapped into a shared latent space. This is augmented by context injection ($W$), where a ``world vector" representing the specific physics or constraints of the current domain is fused with the latent representation.

\subsection{The Objective: Next Primitive Prediction (NPP).} The core of this proposal is a composer architecture (e.g. a transformer-based graph neural network) trained to predict the next algorithmic step and its parameters rather than just emitting surface strings. 
Instead of the next token, the architecture is trained to predict the next node in the PTG:
\begin{equation}
    \mathcal{L}_{NPP} = -\log P(p_{t+1}, c_{t+1} \mid p_{1:t}, c_{1:t}, W)
\end{equation}
here $W$ is the context of the ``possible world.'' The objective forces the model to prioritize a compact, role-flexible basis. Here $p_{t+1}$ represents the predicted Primitive and $c_{t+1}$ represents the predicted Composition Operator. To enforce parsimony, contrastive training can be used to push the model toward learning a minimal basis \cite{chen2020simple, schmidhuber1997discovering}. The model would be penalized if it predicts a complex, non-reusable black-box skill, when a combination of existing primitives could have achieved the same result, similar to DreamCoder \cite{ellis2021dreamcoder}. This forces the model toward compositional parsimony.

We have described parsimony as a minimum-description-length pressure toward the smallest primitive set that reconstructs successful traces; this pressure admits a tractable approximation. During training, a contrastive objective can penalize the model whenever it invokes a complex, non-reusable operation in place of a composition of existing primitives, in the spirit of wake–sleep library learning \cite{ellis2021dreamcoder}. Sparse, steerable directions are already known to exist in trained transformers \cite{todd2024function, lippl2025primitives}, so the geometry such an objective exploits is partly present before fine-tuning rather than built from scratch. Whether this is ultimately cheaper or more expensive than next-token training is an open empirical question.

\begin{proposition}[Compositional pressure toward a minimal basis] If training enforces a parsimony constraint favoring reuse of a minimal primitive set, then optimizing $\mathcal{L}_{NPP}$ biases the system toward representing solutions as compositions over a shared basis $(\mathcal{P}, \mathcal{C})$, rather than as task-specific sequences. Consequently, learned primitives and motifs admit reuse and recombination across distinct task families across possible worlds. 
\end{proposition}

This distinguishes NPP from next-token or next-state prediction \cite{assran2023ijepa, TeohEtAl2025NextLatentPrediction}, which optimize for predictive accuracy over sequences but do not explicitly impose a constraint to push the model to learn a reusable compositional basis.

By making minimal set and rule learning architectural, this framework enables \textit{natively sample-efficient, process-oriented learning}. The system addresses the primitive selection problem by inferring which primitives from a latent library are relevant to a new environment's functional role, rather than relying on superficial similarity to prior episodes. This allows the agent to navigate novel world families by re-binding stable motifs to new environmental settings, a mechanism we term Transfer-as-Recomposition (TaR). In a standard system, if you move from Minecraft to a Voxel World with different gravity, the skills break. In this framework, the NPP realizes that while the values in the state vector have changed, the motif (e.g., Verification $\rightarrow$ Action $\rightarrow$ Verification) is still the optimal path to the goal. It simply re-binds the "Jump" primitive to the new gravity constant. This multiple realizability \cite{putnam1967psychological} of primitives yields generalization.

Finally, the architecture treats composition as a hierarchical, callable, and verifiable process. Discovered motifs, i.e. recurring subgraphs identified through mining successful transition graphs, are "wrapped" and added back to the inventory as higher-order primitives. These compositions can be inspected as graphs or Abstract Syntax Trees (ASTs) \cite{aho2006compilers}, governed by constraints or typing that dictate valid combinations. By integrating verification primitives (e.g., for checking, critiquing, or proving) directly into the library, the architecture ensures that its expanding closure remains grounded in functional, reusable logic that can be shared across \textbf{collective multi-agent settings}.

\subsection{Self-improvement and Self-play in Imagined Worlds}
A natural setting for NPP is an agent that authors its own curriculum: rather than receiving a fixed distribution of environments, it generates \emph{imagined worlds}, counterfactual variants of those it has already mastered, and plays against itself within them. In our terms, such self-play is search not over policies but over the closure $\mathcal{L}(P,C)$. Each generated world is a neighborhood of possible worlds \cite{leibniz1710theodicy, stalnaker1968theory, lewis1986plurality, kripke1980naming} in which the agent varies dynamics, swaps reward structure, or reassigns which objects play which roles, and then asks which primitives and transition motifs still compose into solutions. This recasts the autocurriculum loop \cite{team2021open, wang2019poet}: instead of accumulating ever harder behaviors, the agent expands its closure by discovering primitives that unlock new compositions and motifs that let old primitives be re-bound in qualitatively new ways. 

Crucially, the parsimony constraint is what makes self-play \emph{self-improving} rather than \emph{self-confirming}. Without a minimum-description-length pressure toward a minimal generating set, a self-generated curriculum tends to inflate the library with brittle, world-specific fragments—the compositional analogue of mistaking noise for novelty \cite{hughes2024openendedness}. With it, a primitive earns its place only if it survives recomposition across the agent's own imagined variations, echoing the epistemic move of using a neighborhood of nearby worlds to separate genuine knowledge from a lucky guess \cite{nozick1981philosophical, pritchard2005epistemic}. 

Self-play thus becomes a generator of controlled counterfactuals, and Transfer-as-Recomposition its intrinsic objective: the agent improves not by maximizing return in any single world, but by compressing its successful primitive transition graphs into a smaller basis whose closure covers a wider family of worlds. Verification primitives keep the loop grounded because each recomposed solution is checked before it is wrapped into a higher-order primitive \cite{ellis2021dreamcoder}. As a result, the expanding closure remains a library of reusable components and a grammar to compose them rather than an accumulating record of what once worked.

\subsection{Collective Compositionality and Multi-agent Architectures}
Open-ended exploration and discovery operate best in cumulative multi-agent settings \cite{nisioti2022sapiens, nisioti2024collective}. To support this, architectures must make partial solutions 
transferable. Primitives and compositions should be represented as portable objects retrievable by their functional role, allowing them to be shared and recombined across agents, modules, 
and time. This portability establishes specific architectural requirements, such as the ability to learn from 
observation by decomposing others' behavior into inferred primitives and strategies. 
Furthermore, heterogeneous agents can achieve algorithmic interoperability by developing overlapping 
primitive inventories and sharing a common compositional language or symbolic interface. This framework 
ensures that collective compositions remain adaptive and robust even when the environment changes or 
the team structure is modified.

\section{Evaluation Across Task Families}

Our framework shifts evaluation of open-ended intelligence from behavioral performance to \textit{what persists across possible worlds}. If a capability disappears under a small perturbation, it was likely contingent on original environment rather than a generalized primitive. Open-endedness is thus competence that survives and recomposes across diverse families of environments and tasks, providing a testable meaning to ``general" intelligence as the set of primitives $P$ and composition rules $C$ that remain stable across controlled counterfactual variation.

Evaluation metrics for compositional open-ended intelligence are needed to test whether a system is acquiring a formal grammar of computation rather than accumulating brittle behavioral fragments. For instance, a Primitive Reuse Index (PRI) could measure the frequency of learned primitives $p \in P$ appearing across distinct task families, where a high index identifies general axioms and a low index suggests brittle, task-specific fragments. PRI can help distinguish between memorization and the induction of a minimal generating set. Moreover, a primitive's causal role can be evaluated through \textit{causal isolation and induction} in the residual stream, where ablation of a primitive vector $p$ causes predicted failures, and patching across contexts leads to a predictable transfer of function.

Program scalability can be evaluated via Compositional Depth Generalization(CDG), testing success on Primitive Transition Graphs (PTG) of depth $d_{test} > d_{train}$ to ensure mastery of the closure $\mathcal{L}(P,C)$. A Primitive Discovery Yield (PDY) measure can track the rate at which newly discovered primitives improve utility and sample efficiency. Here, enforcing a parsimony constraint can prioritize motifs that simplify complex trajectories into reusable functions. An Open-ended Curriculum Robustness (OCR) measure can evaluate robustness of primitive composition in producing solutions to shifted task distributions. Finally, a measure of Transfer-as-Recomposition (TaR) can evaluate role-flexibility, holding the primitive library fixed while changing environmental constants or compositional requirements. Low TaR would suggest primitives were likely memorized.

\section{Case Studies}

The case studies that follow ground constraints on the mathematical object proposed above. Physics and mathematics provide the cleanest illustration of minimal generating sets and closure under inference, while neuroscience, collective intelligence, and evolution across species provide existence proof that long-horizon novelty can arise from recombination of reusable building blocks under constraints. Case studies that follow motivate why primitives, their composition rules, and closure rather than raw behavioral repertoires are the appropriate level of description for open-ended intelligence.

\subsection{Case Study: Laws of Physics}
Physics exemplifies the search for a parsimonious minimal generating set of entities and laws that describes an unbounded variety of physical phenomena. Physical reality can be modeled as a formal system derived from a compact physical basis $(P, C)$. This includes a primitive repertoire ($P$), i.e., a finite set of representational primitives, including fundamental constants (e.g., $c, G, \hbar$) and entities like fields or particles, with composition operators ($C$) and algorithmic motifs, such as the principle of least action, symmetry transformations, and conservation laws.
From mechanics to stellar dynamics, a minimal primitive library exhibits open-endedness by generating solutions across unbounded settings. For instance, composing \textit{mass} with \textit{inverse-square attraction} can derive planetary motion through recurrence motifs. Here, the closure $\mathcal{L}(P, C)$ implies that complex phenomena like fluid dynamics and thermodynamics can be derived from the statistical composition of many-body primitives, and a fixed set of equations (e.g., Maxwell's equations) can describe an infinite variety of electromagnetic configurations. Physical process and derivations can be formalized in terms of a primitive transition graph $G = (V, E)$, where nodes ($V$) are instances of physical primitives (e.g., force vectors, states) invoked during a derivation, and edges ($E$) are composition operators (e.g., differential equation steps) transforming system states. Recurring subgraphs, thus, would be compositional motifs (e.g., harmonic oscillations) representing learned rules that transfer across physical domains.
Progress in physics requires the compression and generalization of specific observations into unified field theories. Such compositional generalization enables sample-efficient prediction. 

\subsection{Case Study: Neuroscience} The geometry of neural representations has emerged as a framework for understanding abstraction in computational neuroscience \cite{bernardi2020geometry, chung2021neural, Momennejad2026COBS}. It has been shown that high-dimensional representations with mixed selectivity enable flexible readouts \cite{rigotti2013importance}, providing the computational substrate for compositional reuse. Critically, the prefrontal cortex appears to serve as a recursive composition engine that flexibly combines task-relevant variables to construct solutions \cite{miller2001integrative, duncan2001integrated, rigotti2013importance}, while the hippocampus supports rapid learning of novel associations and relational structures \cite{eichenbaum2017prefrontal, zeithamova2012hippocampus}, and lateral temporal cortices capture compositional semantic motifs and integrated features. This division of labor suggests that compositional intelligence emerges from the coordination of specialized subsystems, each contributing distinct primitives and operations to the overall computation \cite{fuster2001prefrontal, preston2013interplay}. Modeling monkey electrophysiology data shows that the prefrontal cortex and hippocampus represent abstract variables in disentangled geometries \cite{bernardi2020geometry}. 

When humans perform an inferential reasoning task with latent contexts \cite{courellis2024abstract}, hippocampal representations were shown to encode context and stimulus identity along orthogonal directions (disentangled), such that this geometry supports cross-condition generalization (a linear decoder trained in one context generalized to another). Strikingly, this abstract format emerged rapidly, sometimes within minutes following verbal instructions, and was absent on error trials. This demonstrates that the brain actively constructs compositional geometries when inference demands it. A key finding was that parallelism of coding directions, i.e., the degree to which the neural coding vector for a variable (e.g., stimulus A vs. B) is consistent across contexts, increases dramatically during successful inference \cite{courellis2024abstract}. Abstract and parallel coding structures have been extensively documented in prefrontal cortex during flexible task performance and inferential reasoning \cite{rigotti2013importance, bernardi2020geometry, mante2013context}.  In our terminology, parallelism is the degree to which the neural coding vector for a variable or operation is consistent across contexts, and a geometric signature of compositional reuse. 

Brains may solve a version of the minimal generating set problem. The prefrontal cortex may generalize and implement the compositional grammar that combines these primitives into task-specific solutions \cite{miller2001integrative, koechlin2003architecture}, while hippocampal-cortical interactions support learning and consolidation of new primitives into long-term memory \cite{kumaran2016learning, mcclelland1995complementary}. Posterior cortical areas contribute domain-specific representations that serve as the content over which prefrontal composition operates \cite{fedorenko2024language, kanwisher2010functional}.

Neuroscience provides existence proof that compositional structure is an observable property of biological intelligence that we can mathematically formalize. These findings expose a critical gap in current open-ended methods. Systems like POET \cite{wang2019poet}, XLand \cite{team2021open}, and even JEPA \cite{assran2023ijepa} do not explicitly optimize for disentangled or parallel representational geometries. They may incidentally learn such structure, but lack architectural pressure toward the kind of organized primitive inventories that biological systems construct.

\subsection{Case Study: Collective Open-Ended Intelligence} In natural systems, novelty and adaptation often emerge from collective composition \cite{coman2016mnemonic, momennejad2019bridge, couzin2002collective, momennejad2021collective} where multiple agents exchange partial solutions—subroutines \cite{nisioti2024collective}, signals, and constraints—and assemble them into higher-order competence. Example include collectives of animals from schools of fish \cite{couzin2002collective} to wild baboons \cite{strandburg2015shared}. In the primitive-first view, this means that open-ended intelligence is not only the growth of an individual library of primitives, but also the circulation, recombination, and stabilization of primitives across a population. Thus, agents can transmit partial solutions to each other, such that a collective can accumulate these primitives over time and recompose them into new primitives and larger compositions. In this sense, open-ended discovery is a collective dynamic in which intermediate structures circulate, persist, and compound. This is how we can think of scientific practice too, a distributed system that stores partial results (methods, representations, proofs, instruments, protocols), passes them between agents, and continually recombines them into new capabilities.

This logic is visible across biological scales. At the cellular level, individual cells coordinate through signaling, gene-regulatory interactions, and shared molecular pathways that function as reusable micro-primitives (e.g., conditional responses, feedback motifs) that can be deployed in different physiological contexts \cite{alon2007network, shenorr2002network, alon2006introduction}. 

In brains, cognition is distributed across circuits such that while local computations can be specialized, coherent cognition and behavior (e.g., memory recall) arises when these computations are stitched together through communication and gating, yielding compositions that no single sub-circuit can fully accomplish in isolation \cite{mejias2022mechanisms}. At the ecological scale, adaptation and resilience frequently depend on multi-species collaboration, symbioses, microbiomes, collective foraging, and niche construction \cite{odlingsmee2003niche, webster2014cooperation, mcfall2013animals, bordenstein2015host}. Survival is neither individual nor species-local, but a distributed composition of complementary capacities. For artificial multi-agent systems, this shifts objectives from a group maximizing reward to exchanging and recombining partial programs so that (i) discovered subroutines propagate, (ii) heterogeneous agents specialize, and (iii) compositions are robust to changes in team structure or environment. This frames collective open-endedness as a measurable property of compositional exchange.

\subsection{Case Study: Evolutionary Biology}
Evolution can be understood in terms of a compositional process over inherited primitives, and cross-species diversity as evidence for multiple primitive bases or phenotypes. A few more examples of collective open-ended intelligence from nature and evolutionary biology are notable.

\textbf{Genetic ``Toolkits'' (Evo-Devo).} The Hox genes are a perfect example of a minimal set of ``algorithmic primitives'' \cite{carroll2005endless,carroll2005dna}. These same genes control the body plan of a fly, a mouse, and a human \cite{pick2016hox}. The primitives are nearly identical; the composition (timing and location of expression) creates the open-ended diversity of life.

\textbf{Metabolic Pathways.} At the cellular level, life uses a core set of modular chemical reactions, like the Krebs cycle \cite{krebs1937role,nelson2021lehninger} that are recombined to allow organisms to survive in environments ranging from deep-sea vents to arctic tundras \cite{muto2013modular,peregrinvarez2009conservation}.

\textbf{The Immune System.} This is perhaps the best example of ``next primitive prediction.'' The V(D)J recombination process uses a finite set of gene segments (primitives) to compose an essentially infinite variety of antibodies to meet ``problems'' (pathogens) the organism has never encountered before \cite{tonegawa1983somatic,schatz2011recombination}.

\section{Discussion}

\subsection{Distinction from options, HRL, program learning, and library learning} 
Algorithmic primitives differ from the reusable units of earlier frameworks in what binds them. The options framework defines a temporally extended action (or skill) as a tuple consisting of an initiation set, a policy, and a termination condition \cite{sutton1999between, precup2000temporal, precup1998multitime}. Options are typically tied to the environment in which they are acquired and carry no requirement of minimality or transfer. Hierarchical reinforcement learning sequences these skills but leaves them opaque and context-bound, and eigenoptions \cite{machado2017laplacian}, though closer in spirit, supply a spectral basis for a single environment's topology rather than a grammar for composition across environments. Program-synthesis and library-learning systems, DreamCoder \cite{ellis2021dreamcoder}, Stitch \cite{bowers2023stitch}, are the most direct predecessors: they grow a shared inventory by compressing successful traces under a parsimony pressure we share. The decisive difference is the generalization target. Library learning asks whether a shared library better compresses programs for the same task; we ask whether a primitive recovered in one task family reappears, in the same functional role, in a structurally unrelated one. That is a transfer question, not a compression question, and it is what the Primitive Reuse Index and Transfer-as-Recomposition are designed to measure. The distinction is rooted in empirical findings of recent research \cite{lippl2025primitives}, where the authors extract algorithmic primitives (like recall or get-nearest-neighbor-city) from the latent space of an LLM solving a traveling-salesperson task. They then inject the primitive vector while the model is solving another task (e.g., AIME) showing that the response reliably expresses the algorithmic primitive.

\subsection{What Next Primitive Prediction is and is not} 
We specify Next Primitive Prediction at three levels of description. Computationally, NPP predicts which primitive, subgraph, or motif should follow in a reasoning trace, under a description-length constraint favoring the smallest basis able to reconstruct successful traces. Algorithmically, prediction ranges over a structured vocabulary of primitives and composition operators rather than over surface tokens. In implementation, this can be realized as a single end-to-end network, e.g., a transformer-based graph predictor fine-tuned from a pretrained model rather than a hand-assembled collection of modules. References to the residual stream therefore denote emergent operations within one architecture \cite{lippl2025primitives}, not discrete software components. Modular organization need not be engineered in to appear: networks trained on many tasks develop clustered, functionally specialized structure with no architectural prior toward modularity \cite{yang2019task, driscoll2024flexible, johnston2023abstract}. Future research, however, can extend NPP into a teacher-learner or other multi-agent or brain-inspired agentic architectures \cite{webb2025brain}. The present framework is agnostic to the backbone, and a latent-predictive architecture such as JEPA could in principle serve the same role, an option we return to below. 

\textbf{Relation to JEPA} Our objective is also distinct from joint-embedding predictive architectures \cite{assran2023ijepa}, which predict the latent representations of future states and so target state dynamics and representational primitives. JEPA does not, in its current form, press the model toward a minimal inventory of computational operations or toward the transition structure organizing them in multi-step reasoning. It could nonetheless serve as a backbone for NPP: predicting the next primitive rather than the next latent state would require a minimal-set primitive library, a graph-structured prediction target, and a parsimony constraint; additions substantial enough that the result is better understood as a distinct architecture. A direct comparison of JEPA-based and transformer-based instantiations of NPP would be informative for the broader goal of small, sample-efficient models that reason compositionally.

\subsection{The Compositionality Spectrum} We do not claim that all intelligent computation is compositional. Both biological and artificial systems appear to occupy a spectrum from memorization, through cached inference, to full compositional generalization. Episodic retrieval sits near the memorization end; cached structures such as the successor representation \cite{dayan1993successor, momennejad2017successor} occupy the middle, supporting fast statistical inference; and compositional generalization, associated with prefrontal computation and its interaction with the hippocampus, lies at the far end. Cognition is often modeled as trading these modes off, with habitual responses faster but less flexible and compositional inference slower but more transferable \cite{kool2017costbenefit,momennejad2020learning}. We do not claim that composition subsumes the others, but that it is an essential and underdeveloped mode for open-ended intelligence, and that the field currently lacks the formal vocabulary to study it directly.

\subsection{Curriculum Learning, Self-play, and Self-improvement} Curriculum learning \cite{bengio2009curriculum, narvekar2020curriculum, samvelyan2023maestro} and self-play provide a natural setting for compositional open-ended intelligence, but they also expose the limitation of behavioral accounts of progress. In standard autocurricula, improvement is measured by the production of harder tasks, stronger policies, or increasingly competent agents \cite{bengio2009curriculum, narvekar2020curriculum, wang2019poet, wang2020enhanced, team2021open, bauer2023human}. In our framework, the more fundamental question is what changes in the agent's compositional basis. A curriculum is open-ended in the relevant sense when it does not merely generate harder episodes, but induces the discovery, compression, and reuse of primitives and transition motifs whose closure expands across task families. This reframes self-play as a mechanism for generating controlled counterfactuals. When an agent plays against itself or against variants of its own past policies\cite{parkerholder2022accel, samvelyan2023maestro}, it samples nearby possible worlds: altered dynamics, adversarial configurations, new goals, or role reversals. The value of these generated worlds is not only that they improve performance, but that they test which primitives remain useful when surface conditions change. A self-improving system should therefore be evaluated by whether its curriculum expands $\mathcal{L}(P,C)$ while preserving a compact generating set. Without such a parsimony pressure, self-play risks accumulating brittle policies or task-specific hacks. With it, self-play becomes a process for discovering which internal operations survive recomposition and can be promoted into higher-order primitives.

\subsection{The Road Ahead} Progress on open-ended intelligence requires shifting ML's primary objects of study from behavioral and task distributions to \emph{primitives}, \emph{composition operators}, and the \emph{closure} they induce. This yields a concrete research agenda. First, we need shared definitions and measurement protocols for representational primitives, algorithmic primitives, compositional motifs, and closure. These objects should be measurable across models, tasks, and training regimes, rather than introduced only as post hoc interpretations. 

Second, we need a mathematics of curriculum and self-improvement at the level of compositional structure. Current autocurricula and self-play systems generate increasingly difficult tasks or stronger policies, but they do not directly measure whether the agent is discovering a smaller and more reusable basis. In the present framework, a curriculum should be evaluated by whether it expands $\mathcal{L}(P,C)$ while preserving or compressing the primitive set. This motivates metrics for tracking when new primitives emerge, when motifs consolidate, and when old primitives are re-bound to new roles across counterfactual task families. 

Third, future architectures should make this process explicit. Next Primitive Prediction is one proposal: rather than learning only to predict tokens, states, or latent features, a model is trained to predict the next primitive or motif in a reasoning trace under a parsimony constraint. A self-improving version of such a system would generate its own task variations, extract primitive transition graphs from successful solutions, test whether the same primitives recombine across worlds, and consolidate the motifs that survive. Self-play then becomes more than competition or data generation; it becomes a mechanism for expanding and compressing the agent's compositional basis. 

Fourth, we need comparative empirical work across neuroscience, evolution, and ML. Biological systems provide evidence that flexible cognition depends on structured reuse, mixed selectivity, compositional geometry, and distributed coordination. Artificial systems now make it possible to test analogous questions mechanistically: when do primitives emerge, what objectives induce them, how stable are they across model scale and fine-tuning, and which transition motifs predict transfer? These measures would complement behavioral evaluation by exposing the internal compositional strategies that determine whether a system can generalize to new families of worlds. 

Finally, open-ended discovery is often cumulative and collective. Future work should study agents that communicate primitives, infer the compositional strategies of others, and recombine partial solutions across individuals and time. A mature theory of open-ended intelligence should therefore connect individual learning, self-improvement, and collective accumulation under a single formal object: the growth and compression of compositional closure.

\section{Alternative Views}

One potential critique arises from the original open-ended framework \cite{stanley2017openended}, which advocates for the abandonment of objectives in favor of novelty search. However, without a focus on primitives, this approach fails to produce replicable 'stepping stones', making it difficult to guarantee that specific solutions can be reconstructed across different evolutionary runs \cite{stanley2015greatness}. Another critique could come from the 'Reward is Enough' hypothesis \cite{silver2021reward}, which posits that intelligence emerges solely from maximizing complex rewards, viewing explicit mathematical objects like primitives as constraints that might limit a system's ultimate ceiling. Similarly, proponents of the 'Bitter Lesson' \cite{sutton2019bitter}, which argues that massive scaling of computation and data is more effective than architectural inductive biases, could object that systematic generalization is merely an emergent byproduct of scale. 

We address these views by noting that while scale and reward drive raw performance, they do not inherently provide the mechanistic interpretability or sample efficiency required for agents to safely adapt to data-scarce "world families". Finally, our work diverges from the case of eigen-options in hierarchical RL \cite{machado2017laplacian}, where the primary bottleneck was identifying relevant options among a vast set. Unlike brittle, environment-dependent eigen-options, our framework centers on a minimal transferable set of algorithmic primitives. By identifying these common motifs across architectures, we can automate their discovery and ensure their emergence even within novel systems.

\section{Conclusion} 
Our framework changes the unit of analysis for open-ended intelligence from behaviors and tasks to primitives, composition operators, a compositional grammar, and the closure they induce. This shift unifies explanatory and architectural programs, because the objects that make an architecture interpretable also become the objects that can be learned and optimized. As a result, algorithmic interpretability is no longer merely a post hoc analysis of fixed models, but a source of inductive biases and training objectives that shape both the pretraining and post-training of new ones.  This establishes a concrete research and development program to identify primitives and motifs, test whether they persist and recombine across worlds, and use those that do to define the objectives and structure of open-ended and continual learning systems.

\bibliography{refs}
\bibliographystyle{icml2026}

\end{document}